\documentclass[letterpaper, 10 pt, conference]{ieeeconf}  

\IEEEoverridecommandlockouts                              

\usepackage{graphicx} 
\usepackage{subcaption}

\usepackage{xcolor}
\usepackage{amsmath, xparse}
\usepackage{amsfonts}
\usepackage{amssymb}
\usepackage{hyperref}
\usepackage{siunitx}
\usepackage{multirow}

\newcommand{\apriltag}[1]{AprilTag{#1}}
\newcommand{\modelname}[1]{TartanCalib{#1}}

\newcommand{\mat}[1]{\mathbf{#1}}
\newcommand{\vct}[1]{\mathbf{#1}}
\newcommand{\ssymb}[1]{\boldsymbol #1}
\newcommand{\fsymb}[1]{\boldsymbol #1}

\newcommand{\tsp}{\mathrm{\intercal}}
\newcommand{\st}[1]{\mathrm{#1}}

\newcommand{\intextfunc}[1]{\textit{#1}}
\renewcommand{\baselinestretch}{0.947}
\title{\LARGE \bf
\modelname{}: Iterative Wide-Angle Lens Calibration \\ using Adaptive SubPixel Refinement of \apriltag{s}
}

\author{Bardienus P. Duisterhof$^{1}$~~~Yaoyu Hu$^{1}$~~~Si Heng Teng$^{1}$~~~ Michael Kaess$^{1}$ ~~~Sebastian Scherer$^{1}$
\thanks{$^{1}$The Robotics Institute, Carnegie Mellon University,
Pittsburgh, PA 15213, USA.  \url{{bduister,yaoyuh,sihengt,kaess,basti}@andrew.cmu.edu}}
}

\begin{document}

\maketitle
\thispagestyle{empty}
\pagestyle{empty}

\begin{abstract}

Wide-angle cameras are uniquely positioned for mobile robots, by virtue of the rich information they provide in a small, light, and cost-effective form factor. An accurate calibration of the intrinsics and extrinsics is a critical prerequisite for using the edge of a wide-angle lens for depth perception and odometry. Calibrating wide-angle lenses with current state-of-the-art techniques yields poor results due to extreme distortion at the edge, as most algorithms assume a lens with low to medium distortion closer to a pinhole projection. In this work we present our methodology for accurate wide-angle calibration. Our pipeline generates an intermediate model, and leverages it to iteratively improve feature detection and eventually the camera parameters. We test three key methods to utilize intermediate camera models: (1) undistorting the image into virtual pinhole cameras, (2) reprojecting the target into the image frame, and (3) adaptive subpixel refinement. Combining adaptive subpixel refinement and feature reprojection significantly improves reprojection errors by up to \SI{26.59}{\percent}, helps us detect up to \SI{42.01}{\percent} more features, and improves performance in the downstream task of dense depth mapping. Finally, \modelname{} is open-source and implemented into an easy-to-use calibration toolbox. We also provide a translation layer with other state-of-the-art works, which allows for regressing generic models with thousands of parameters or using a more robust solver. To this end, \modelname{} is the tool of choice for wide-angle calibration. Project website and code: \url{http://tartancalib.com}.

\end{abstract}

\section{Introduction}

Cameras with wide-angle lenses enlarge the field-of-view (FOV) of a mobile robot in a compact form factor. This feature provides many benefits for crucial tasks such as visual odometry and depth mapping since a larger FOV brings more visual features. However, careful calibration is mandatory to fully leverage the image regions with high levels of distortion. Typically, calibration is obtained by showing the camera a known calibration target, which is then used to estimate the intrinsics of the camera model and extrinsics for multi-camera systems. Unlike ordinary lenses, wide-angle lenses, due to the high level of distortion in the image, impose certain special challenges: (1) it is challenging to robustly and accurately detect the visual features of the calibration target, due to extreme lens distortion. (2) The camera models available may not fit the lens very well. Previous work predominantly focused on obtaining more suitable camera models~\cite{9156397,usenko18double-sphere,Devernay2001StraightLH,1642666}, or finding better procedures to fit the camera models~\cite{9711344,Scaramuzza2006AFT,4059340,URBAN201572}. Consequentially, most publicly available tools only provide acceptable calibration results in the middle region of a wide-angle lens, where distortion is moderate, leaving the majority of the highly distorted border areas unusable. To make more image regions usable and effectively explore the visual information embedded near the border of a wide-angle lens, we need better calibration that delivers improved intrinsics and extrinsics parameter estimation.

\begin{figure}
    \centering
    \includegraphics[width=\columnwidth]{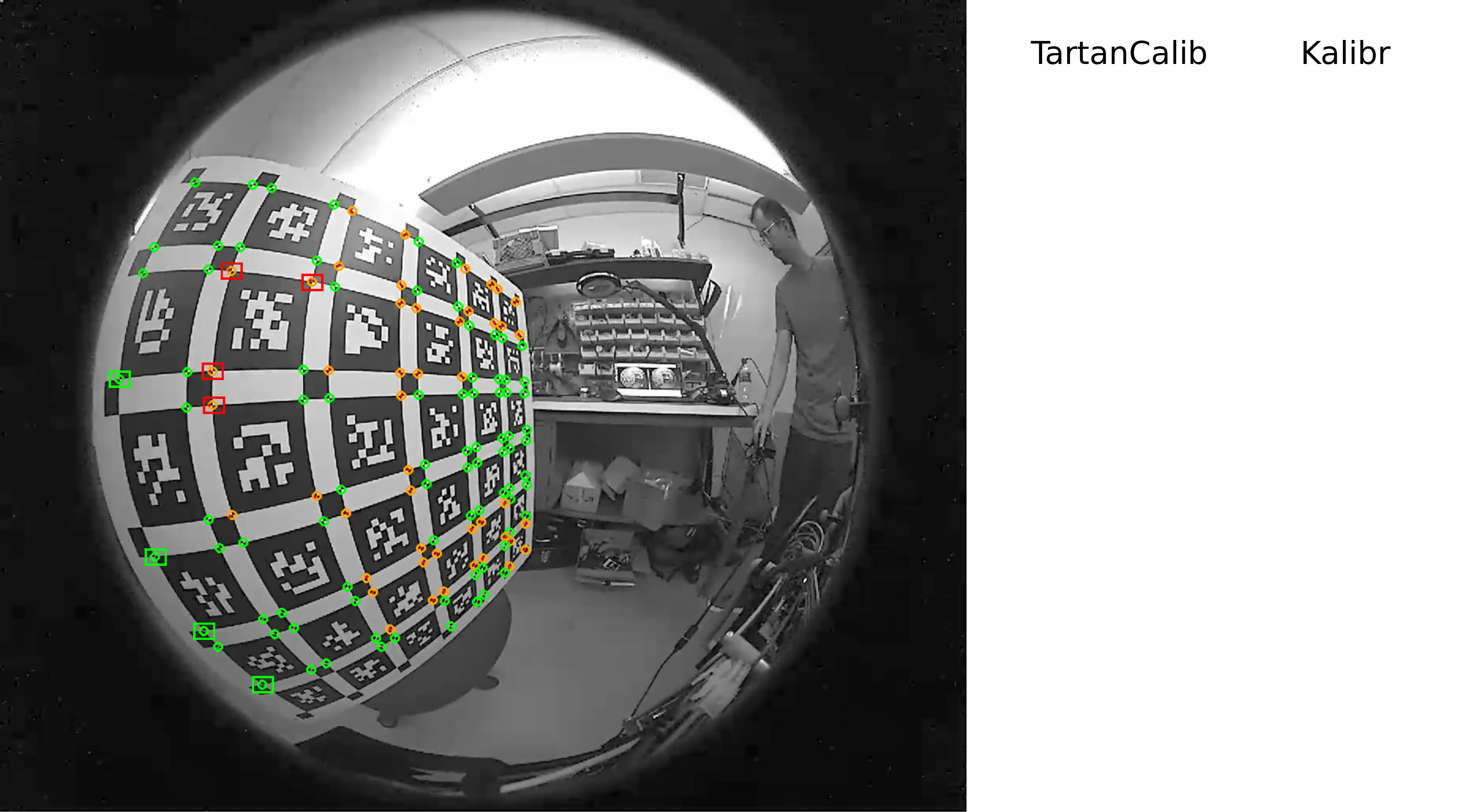}
    \caption{Comparison of target detection and feature refinement between \modelname{} and Kalibr.
    (Left) Target detection. Green circles: features newly detected by \modelname{}, orange circles: features previously picked up by Kalibr~\cite{kalibr}. (Right) Zoomed-in view of detected features. Green point inside green and orange circles: refined features of \modelname{}, red point inside orange circle: features from Kalibr. \modelname{} detects more features near the image border and the features have better location accuracy.}
    \label{fig:reprojection}
    \vspace{-5mm}
\end{figure}

Here, we focus on accurate and robust target detection in the high distortion regions. State-of-the-art target calibration pipelines~\cite{5979561,8967787,9156397} fail in presence of high distortion, as shown in Figure~\ref{fig:reprojection}. The two key procedures of these calibration pipelines, target detection and feature refinement, rely on the assumption of low to medium distortion or a camera projection that is close to a pinhole camera. These assumptions are violated in the case of wide-angle lenses, especially near the image border. The result is that many features are either not detected or are not detected in an accurate way. For those sparsely detected border features, their pixel locations tend to be inaccurate because of poor performance of the feature refinement methods in the high distortion regions. The above issues deteriorate the quality of the estimated camera model and limit the usability of the border region of a wide-angle lens. 


\begin{figure}
    \centering
    \includegraphics[width=\columnwidth]{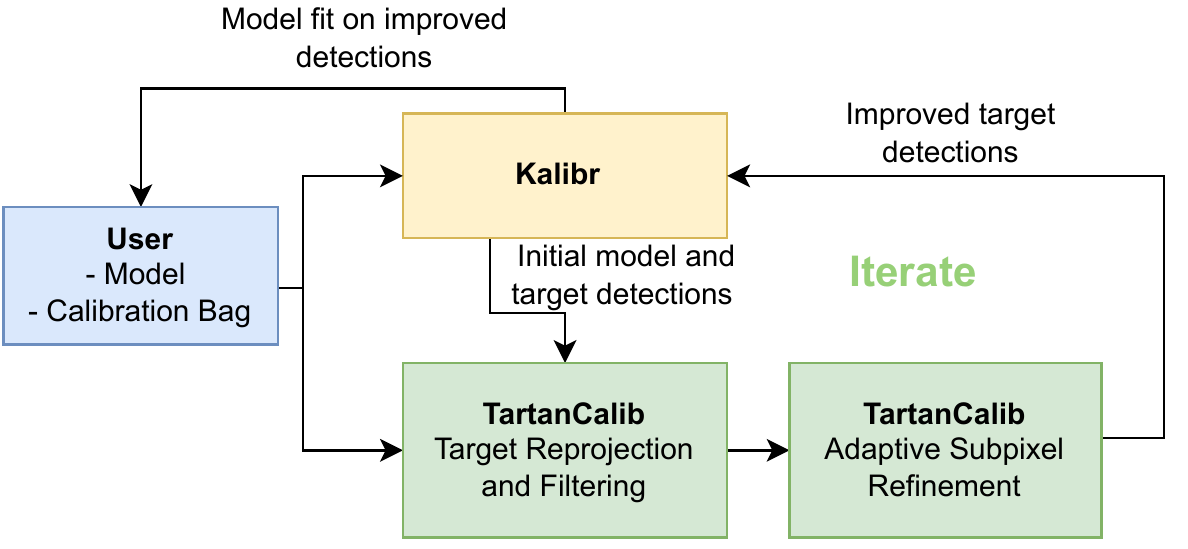}
    \caption{The pipeline of \modelname{}. The pipeline consists of an iterative calibration procedure with a newly proposed adaptive feature refinement method.}
    \label{fig:pipeline}
    \vspace{-5mm}
\end{figure}

In this work, we propose an iterative calibration pipeline (Figure~\ref{fig:pipeline}), which consists of three core elements: (1) undistortion of the original image, (2) reprojecting the target into the image frame using the intermediate camera model, and (3) adaptive subpixel refinement based on reprojected target size. For our method, we develop two new subpixel feature refinement methods to facilitate accurate target detection in highly distorted regions, towards a better overall calibration in the border area of a wide-angle lens. Our contributions are:

\begin{itemize}
  \item A novel methodology for wide-angle lens calibration, using iterative target reprojection and adaptive subpixel refinement.
  \item We show the benefit of our pipeline using traditional quality metrics such as  reprojection error, as well as some downstream tasks relevant to mobile robots.
  \item We present our pipeline as an open-source easy-to-use package, which we term `\modelname{}'.
\end{itemize}


Using our method, we find up to \SI{42.01}{\percent} more features, and up to \SI{26.59}{\percent} lower overall reprojection error. The entire pipeline is made open-source and can be easily integrated into Kalibr.

\section{Related Work}
This Section lays down the related work in the areas of camera models (Section~\ref{sec:related_cam_models}), calibration toolboxes (Section~\ref{sec:related_toolboxes}), and pattern design and feature detection (Section ~\ref{sec:related_pattern}).

\subsection{Camera Models}
\label{sec:related_cam_models}
A typical calibration procedure needs to select a parametric or generic camera model for a lens and estimates the model parameters during the calibration process. There are a number of models designed specifically for wide-angle lenses, as their projection is significantly different from low-distortion cameras. Some of the more common models are the Double Sphere model~\cite{usenko18double-sphere}, the Kannala-Brandt model~\cite{1642666}, and the Field-of-View model~\cite{Devernay2001StraightLH}. Parametric models typically have only a few degrees of freedom, making their parameters easier to be estimated as compared with generic models, but providing a trade-off in accuracy. Generic models have far more parameters, aiming to more accurately represent lens geometry. It has been shown that these models have a significantly lower reprojection error~\cite{9156397}. A distinction can be made between non-central generic models and generic models. Central generic models~\cite{8500466,central_generic_2} assume all observation lines intersect in the center of projection, whereas non-central generic models do not make that assumption~\cite{937611,7516654}. Typically non-central generic models perform better but may be more complicated to deploy (e.g., undistortion to a pinhole image is not possible without knowing pixel depth). Our toolbox, \modelname{}, supports both parametric and generic camera models to achieve the best possible calibration. 

\subsection{Calibration Toolboxes}
\label{sec:related_toolboxes}
As geometric camera calibration is an important prerequisite for many machine vision applications, numerous calibration toolboxes have been developed.~\cite{URBAN201572,10.1007/978-3-540-24671-8_1,888718}. The famous computer vision package OpenCV~\cite{opencv} has its own wide-angle lens calibration support, and supports checkerboard targets. OcamCalib~\cite{URBAN201572} is another well-known toolbox, using exclusively (less accurate~\cite{5979561,8237833}) checkerboard targets for calibration. 

Recently BabelCalib~\cite{9711344} was proposed, with its robust optimization strategy being the key advantage. However, the most commonly used calibration toolbox is Kalibr~\cite{kalibr}, which is easy to use and allows for retrieving the intrinsics and extrinsics of multiple cameras with a wide variety of camera models and targets. In this work, \modelname{} is integrated into Kalibr~\cite{kalibr} as an easy-to-use toolbox. In addition to Kalibr, \modelname{} also supports the the use of BabelCalib as a solver, and the generally more accurate generic models~\cite{9156397}.

\subsection{Pattern Design and Feature Detection}
\label{sec:related_pattern}
Target detection is one of the key functions of a calibration pipeline. 
By far the most commonly used calibration targets are checkerboard~\cite{opencv}, dot patterns~\cite{879788}, and \apriltag{s}~\cite{5979561}. Dot patterns are susceptible to perspective and lens distortion, whereas a checkerboard tends to fail when it is only partially observed, which makes calibrating wide-angle lenses extremely challenging. Some researchers proposed to use novel patterns, such as triangle features~\cite{8237833,9156397} to increase the gradient information, but these typically are not robust enough for the high distortion as present at the edge of a wide-angle (fisheye) lens.

In~\cite{9156397}, the authors use a single \apriltag{} to determine the pose of a custom target and assume a homography as a camera model to reproject the target onto the image frame.  This approach has three fundamental issues for high-distortion wide-angle lenses: (1) using a single \apriltag{} is not robust enough, (2) using a homography as a camera model for target reprojection will impose a reprojection error that makes it impossible to recover the true target position, and (3) the refinement method used is shown to be unstable for \apriltag{s} (and checkerboards).

Inspired by~\cite{9156397}, we propose a pipeline designed for a grid of \apriltag{s}. \modelname{} adopts an iterative process, that makes it possible to use a relatively accurate intermediate camera model instead of a homography, to reproject target points into the image frame. Additionally, we propose two novel adaptive subpixel refinement methods, arriving at more features detected with superior sub-pixel accuracy. 

\section{Preliminaries}
This section addresses the preliminaries related to notation (Section~\ref{sec:pre_notation}) and camera models (Section~\ref{sec:pre_cam_models}), necessary for this work.
\subsection{Notation}
\label{sec:pre_notation}
Our notations are inspired by \cite{usenko18double-sphere}.
In the equations presented later, lowercase characters are scalars (e.g., $\alpha$), whereas lowercase bold characters (e.g., $\vct{x}$) are vectors. Matrices are denoted using bold uppercase letters (e.g., $\mat{T})$. 
Pixel coordinates are represented as $\vct{u} = [u,v]^{\tsp} \in 
\ssymb{\Theta} \subset \mathbb{R}^2 $, here $\ssymb{\Theta}$ is the image domain points. 3D points are presented as $\vct{x} = [x,y,z]^{\tsp} \in \ssymb{\Omega} \subset \mathbb{R}^{3}$, here $\ssymb{\Omega}$ is the subset of valid 3D points that can be projected into the image frame. We denote the transformation from the camera frame to the calibration target frame as $\mat{T}_{\st{tar}} \in SE(3)$. The image matrix is denoted as $ I$.

\subsection{Camera Models}
\label{sec:pre_cam_models}

A camera model typically consists of a projection and unprojection function. The projection function $\fsymb{\pi} \, \mathbf{:} \,  \ssymb{\Omega} \rightarrow \ssymb{\Theta}$ projects a 3D point to image coordinates. Its inverse, the unprojection function: $\fsymb{\pi}^{-1} \, \mathbf{:} \, \ssymb{\Theta} \rightarrow \mathbb{S}^2 $ unprojects image coordinates onto a unit sphere. The projection of a 3D point can be described as $\fsymb{\pi}(\vct{x},\vct{i})$ where $\vct{x}$ is a point in 3D space, and $\vct{i}$ is the set of parameters for the camera model. Similarly, the unprojection function is denoted as $\fsymb{\pi}^{-1}(\vct{u},\vct{i})$, where $\vct{u}$ is the coordinate in image space.

\section{Method}
The high-level idea behind \modelname{} (Figure~\ref{fig:pipeline}) is to iteratively optimize a camera model, by leveraging intermediate camera models to improve target detection. The iteration includes several key components that will be detailed in the following sections. The components are Undistortion (Section~\ref{sec:method_undistortion}), Target Reprojection (Section~\ref{sec:reprojection}), Corner Filtering (Section~\ref{sec:corner_filtering}), and Subpixel Refinement (Section~\ref{sec:corner_refinement}).

The visual features of the calibration target manifest themselves as corner features. In latter sections, we interchangeably refer to corners as target features.

\subsection{Undistortion}
\label{sec:method_undistortion}
An intuitive way to improve target detection in wide-angle camera calibration is to undistort the image into multiple pinhole reprojections. This approach should get rid of some of the difficulties caused by highly distorted targets. To undistort the image we model a virtual pinhole camera, which has four parameters $\vct{i} = [f_x,f_y,c_x,c_y]^{\tsp} $. The projection function is defined as:
\begin{equation}
\label{eq:pinhole}
\fsymb{\pi} (\mathbf{x},\mathbf{i}) = \begin{bmatrix} 
f_x \frac{x}{z} \\ 
f_y \frac{y}{z}
\end{bmatrix}
+ \begin{bmatrix}
c_x \\
c_y
\end{bmatrix}
\end{equation}
where $f_x$ and $f_y$ are focal length and $c_x$ and $c_y$ are the pixel coordinate of the principle point.
Creating a virtual pinhole camera is possible by first unprojecting the pinhole pixel coordinates to $\mathbb{S}^{2}$ space, to then reproject those points back into the distorted image frame. We then query the pixel location at that location in the distorted frame and substitute it back into the pinhole image to arrive at an undistorted image.

\subsection{Target Reprojection}
\label{sec:reprojection}
 While undistortion reduces lens distortion, we may still be unable to detect the target due to perspective distortion and other visual artefacts such as motion blur. As proposed in~\cite{9156397}, it is possible to reproject known target coordinates back into the image frame without actually detecting the target. The authors show that a homography can be used for this purpose. 
Equation~\ref{eq:reproject_target} shows how a camera model can be used to reproject a point from target coordinates ($\vct{x}_{\st{t}}$) to image coordinates ($\vct{u}_{\st{t}}$).
 
 \begin{equation}
 \label{eq:reproject_target}
     \vct{u} =  \fsymb\pi (\vct{x},\vct{i}) = \fsymb{\pi} (\mat{T}_{\st{tar}}\cdot \vct{x}_{\st{t}},\vct{i})
 \end{equation}
 
 Here $\mat{T}_{\st{tar}}$ is the transformation from the target frame to the camera frame, $\vct{x}_{\st{t}}$ is a coordinate in the target frame, and $\vct{x}$ is a vector in the camera frame.

\subsection{Corner Filtering}
\label{sec:corner_filtering}
While reprojecting the target into the image frame using an intermediate camera model may yield somewhat accurate estimates, it is uncertain if all of the target is visible in the frame. Therefore, a filtering policy is required, only keeping the features (corners) that appear in the frame. We achieve robust filtering by following these steps: 1) loop over all detected quads (detected squares), 2) check if all 4 corners of each quad are close to a reprojected target corner, and 3) perform subpixel refinement on all corners.

\subsection{Subpixel Refinement}
\label{sec:corner_refinement}
Subpixel refinement is required to translate the features reprojected using the intermediate model into features that actually match the corners as seen in the image. We propose two algorithms: 1) a simple modification to OpenCV's \intextfunc{cornerSubPix()} function \cite{cornersubpix}, and 2) a symmetry-based refinement method specifically designed for high-distortion lenses. 

\subsubsection{Adaptive cornerSubPix()}

\intextfunc{cornerSubPix()}~\cite{cornersubpix} computes the image gradient within a search window in order to iteratively converge towards the corner. In doing this, the size of the search window is a critical hyperparameter: if the search window is too small, the algorithm may never find the corner, whereas a large window will yield inaccurate results or even converge to another corner. The window size is typically a fixed parameter, not changed for different image resolutions or distortion levels. Figure~\ref{fig:reprojection} demonstrates the resulting problem when the window size is chosen inappropriately.

In this work we present an adaptive version of \intextfunc{cornerSubPix()}, that changes window size based on the tag appearance in the image frame. Equation~\ref{eq:adaptive_cornersubpix} shows how the size of the resize window is determined. The algorithm reprojects all target features into the image frame, and for each feature finds it nearest neighbor in the image frame.  We then use that information to scale the search window, according to Equation~\ref{eq:adaptive_cornersubpix}.
%
\begin{equation}
\label{eq:adaptive_cornersubpix}
    w_{\mathbf{x}_{\st{t}}} = \min\limits_{\vct{x}^{*}_{\st{t}} \in \ssymb{Q}} s \cdot \left( \fsymb{\pi} ( \mat{T}_{\st{tar}}\cdot \vct{x}_{\st{t}},\vct{i} ) - \fsymb{\pi} (\mat{T}_{\st{tar}} \cdot \vct{x}^{*}_{\st{t}},\vct{i}) \right)
\end{equation}
Here $s$ is a user-defined scalar that can be used to make the window either bigger or smaller in the target frame. $\vct{x^{*}}$ is a target coordinate within $\ssymb{Q}$, the subspace of $\mathbb{R}^{3}$ that consists all possible feature coordinates in the target frame.

\begin{table}[!tb]
\caption{Number of features detected in datasets collected on a GoPro Hero 8 camera and the Lensagon BF5M ultra-wide fisheye lens. The percentages signify the fraction of features detected w.r.t. the theoretical maximum (i.e., number of frames $\times$ number of features)  }
\centering
\begin{tabular}{lll}
\hline
                                   & GoPro Hero 8                       & Lensagon BF5M \\ \hline
Deltille                            & \textbf{70,464 (\SI{97.87}{\percent})}    & 34,149 (\SI{47.43}{\percent})            \\
\apriltag{}3                        & 45,788 (\SI{63.59}{\percent})             & 29,332 (\SI{40.74}{\percent})            \\
\apriltag{} Kaess                   & 68,708 (\SI{95.43}{\percent})             & 38,456 (\SI{53.41}{\percent})            \\
Aruco (OpenCV)                      & 66,092 (\SI{91.79}{\percent})                  & 22,364 (\SI{31.06}{\percent})            \\
Kalibr                              & 62,383 (\SI{86.64}{\percent})                  & 33,433 (\SI{46.43}{\percent})            \\
\modelname{}                        & 67,693 (\SI{94.01}{\percent})                  & \textbf{54,685} (\SI{75.95}{\percent})   \\ \hline
\end{tabular}
\label{tab:features_detected}
\vspace{-5mm}
\end{table}

\subsubsection{Symmetry-Based Refinement}
The authors of~\cite{9156397} first proposed refining calibration target detections by optimizing a symmetry-based cost function. The original cost function is shown in Equation~\ref{eq:symmetry_original}.

\begin{equation}
\label{eq:symmetry_original}
    C_{\st{sym}}(\textbf{H}) = \sum_{k=1}^{n}(I(\mathbf{H}(\vct{s}_k))-I(\mathbf{H}(-\vct{s}_k)))^{2}
\end{equation}

Here $\mathbf{H}$ is a homography from the target frame to the image frame, with the center in the target frame being the feature that is being refined. $\vct{s}_k$ and $-\vct{s}_k$ are two samples in the target frame, that are defined such that the origin corresponds to the feature location. The authors then optimize $\mathbf{H}$ with the Levenberg-Marquardt method in order to minimize $C_{\st{sym}}$. The authors show that using deltille grids, checkerboard, and their custom target, this method is the most precise refinement method. Their custom target consists of a single \apriltag{} to determine a homography between the target plane and image plane, which is surrounded by a number of star-shaped feature points. The homography determined using the single \apriltag{} is used to make an initial guess for the homography for each feature, it is therefore important that the camera being calibrated behaves somewhat close to a pinhole camera.

As a result, this approach is effective for low-distortion to medium-distortion as the projection is close to a homography. For true ultra-wide lenses, however, the proposed approach fails for two reasons: (1) the single \apriltag{} is not detected, and (2) the homography assumption causes the refinement procedure to fail.

We propose an adapted version that optimizes a modified symmetry-based objective function, shown in Equation~\ref{eq:fe_symmetry}.
\begin{equation} \label{eq:fe_symmetry}
\begin{split}
    C_{\st{sym}}(\vct{x}_{\st{t}}) = \sum_{k=1}^{n} ( & I(\fsymb{\pi} (\mat{T}_{\st{tar}} \cdot (\vct{x}_{\st{t}}+\vct{s}_{k}), \vct{i})) \\ 
    - & I(\fsymb{\pi} (\mat{T}_{\st{tar}} \cdot (\vct{x}_{\st{t}}-\vct{s}_{k}), \vct{i}) ) )^{2}
\end{split}
\end{equation}

Here $\vct{x}_{\st{t}}$ is the location of the feature in target space, that is transformed to the camera frame using $\mat{T}_{\st{tar}}$, and projected to the image frame using the projection function $\fsymb{\pi}$. All steps in this equation are differentiable, which we use to optimize the feature location in the target frame directly. With this we leverage the intermediate model to retrieve higher quality symmetric samples, and a better initial estimate.

\begin{table*}[!htb]
\centering
\caption{\label{tab:ord_6x6_distr} The normalized number of features detected using the BF5M ultra-wide fisheye camera, Sorted by polar angle. The results clearly show that \modelname{} detects most features, especially at the edge of the lens.    }

\resizebox{0.9 \linewidth}{!}{\begin{tabular}{lllllllllll}
\hline
\multicolumn{11}{c}{Normalized Number of Features Detected, Sorted by Polar Angle }          \\ \hline
& \SI{0}{\degree}-\SI{10}{\degree}          & \SI{10}{\degree}-\SI{20}{\degree}         & \SI{20}{\degree}-\SI{30}{\degree}         & \SI{30}{\degree}-\SI{40}{\degree}         & \SI{40}{\degree}-\SI{50}{\degree}         & \SI{50}{\degree}-\SI{60}{\degree}         & \SI{60}{\degree}-\SI{70}{\degree}         & \SI{70}{\degree}-\SI{80}{\degree}         & \SI{80}{\degree}-\SI{90}{\degree}         & \SI{90}{\degree}-\SI{100}{\degree}        \\ \hline
Kalibr             & 0.75          & 0.80          & 0.87          & 0.84          & 0.72          & 0.56          & 0.48          & 0.38          & 0.19          & 0.04          \\
Deltille           & 0.62          & 0.61          & 0.63          & 0.63          & 0.64          & 0.64          & 0.63          & 0.60          & 0.52          & 0.60          \\
AT3                & 0.75          & 0.75          & 0.71          & 0.63          & 0.57          & 0.53          & 0.46          & 0.32          & 0.19          & 0.09          \\
Kaess              & 0.78          & 0.83          & 0.92          & 0.91          & 0.81          & 0.68          & 0.63          & 0.50          & 0.24          & 0.05          \\
Aruco              & 0.89          & 0.85          & 0.82          & 0.63          & 0.43          & 0.24          & 0.10          & 0.02          & 0.00          & 0.00          \\
TartanCalib  & \textbf{1.00} & \textbf{1.00} & \textbf{1.00} & \textbf{1.00} & \textbf{1.00} & \textbf{1.00} & \textbf{1.00} & \textbf{1.00} & \textbf{1.00} & \textbf{1.00} \\ \hline
\end{tabular}}
\vspace{-5mm}
\end{table*}
\section{Results}
\subsection{Evaluation Metrics}
In this Section, we evaluate a number of metrics in order to compare \modelname{} against other state-of-the-art approaches for feature detection and geometric camera calibration. Evaluating geometric camera calibration is challenging as no ground truth exists for feature detections and camera geometry. A metric that is traditionally used for geometric camera calibration is reprojection error, especially when different models or optimization routines are compared. 
This is only an appropriate metric when the features are the same, as a different feature distribution yields an entirely different optimization problem. 
We will see that \modelname{} brings more detected features in the entire image, especially in the highly distorted regions, fulfilling its design purpose. Reprojection error could be higher if the camera model is not complex enough to fit all the detected features in the distorted regions. This observation indicates that using reprojection error as the metric is not appropriate.
Nevertheless, the reprojection error is presented for reference, while we use other quantitative metrics to show that the accuracy and coverage of the detected features from \modelname{} are of higher quality.

\begin{figure}[!htb]
    \centering
    \includegraphics[width=0.7\linewidth]{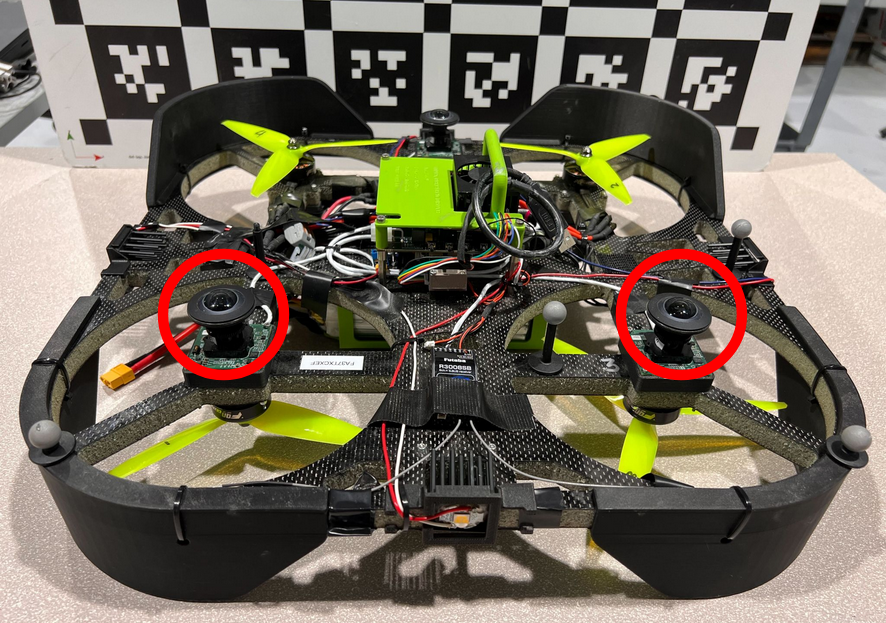}
    \caption{The aerial robot used in the experiment. It has three identical sets of cameras and fisheye lenses. Every camera uses a Lensagon BF5M2023S23C Lens (\SI{195}{\degree} FOV), with a Leopard Imaging LI-IMX264-MIPI-M12 sensor. Red circles: the two fisheye lense that are used in Section~\ref{sec:stereo_depth} to form a stereo pair.}
    \label{fig:drone}
    \vspace{-5mm}
\end{figure}

\subsection{Experimental Setup}
\label{sec:experiments_setup}
To evaluate \modelname{}, its performance is evaluated in two camera configuration: (1) an aerial robot equipped with multiple cameras ultra-wide Lensagon BF5M fisheye lenses (Figure~\ref{fig:drone}, \SI{195}{\degree} FOV), and (2) a GoPro Hero 8 with a \SI{122.6}{\degree} $\times$ \SI{149.2}{\degree} FOV. For the GoPro, data is collected using its proprietary software, whereas the ROS gscam drivers are used to extract images from the cameras on the aerial robot. The Lensagon BF5M fisheye lens is installed on a Leopard Imaging LI-IMX264 camera. This combination will be referred to as the Lensagon BF5M from now on. We use an \apriltag{} target for calibration experiments. The board presents a  6 $\times$ 6 grid with an \SI{8.8}{\cm} tag size. For each experiment, 500 images are retrieved per camera by subsampling each dataset.

\subsection{Feature Detection}
The first metric considered is the number of features detected, and their coverage in spherical coordinates. In this section we consider undistortion and target reprojection as methods for retrieving more features, and compare them against state-of-the-art approaches for \apriltag{} detection.

\subsubsection{Undistortion}
An intuitive way to improve target detection in wide-angle camera calibration is to undistort the image into multiple pinhole projections. This idea is evaluated by iteratively undistorting the image into five pinhole projections. The first plane is pointed in the positive $z$ direction (the principle axis) of the camera frame, while the other projections are pointed at a polar angle of \SI{90}{\degree}. The azimuth angles start from zero and are incremented by \SI{90}{\degree} for each plane. The five planes together form a cube. In correspondence with Figure~\ref{fig:pipeline}, the undistorted pinhole projections are used to iteratively improve the camera model.

To quanitfy the effect of undistortion, the previously collected images for Lensagon BF5M are further sampled into a subset that contains 100 frames. \modelname{} is then used to iteratively undistort the wide-angle images into pinhole projections. In this experiment undistortion yielded an additional $\SI{5}{\percent}$-$\SI{10}{\percent}$ of features.


\subsubsection{Reprojection}

\begin{figure}[!htb]
    \centering
  \includegraphics[width=0.9\linewidth]{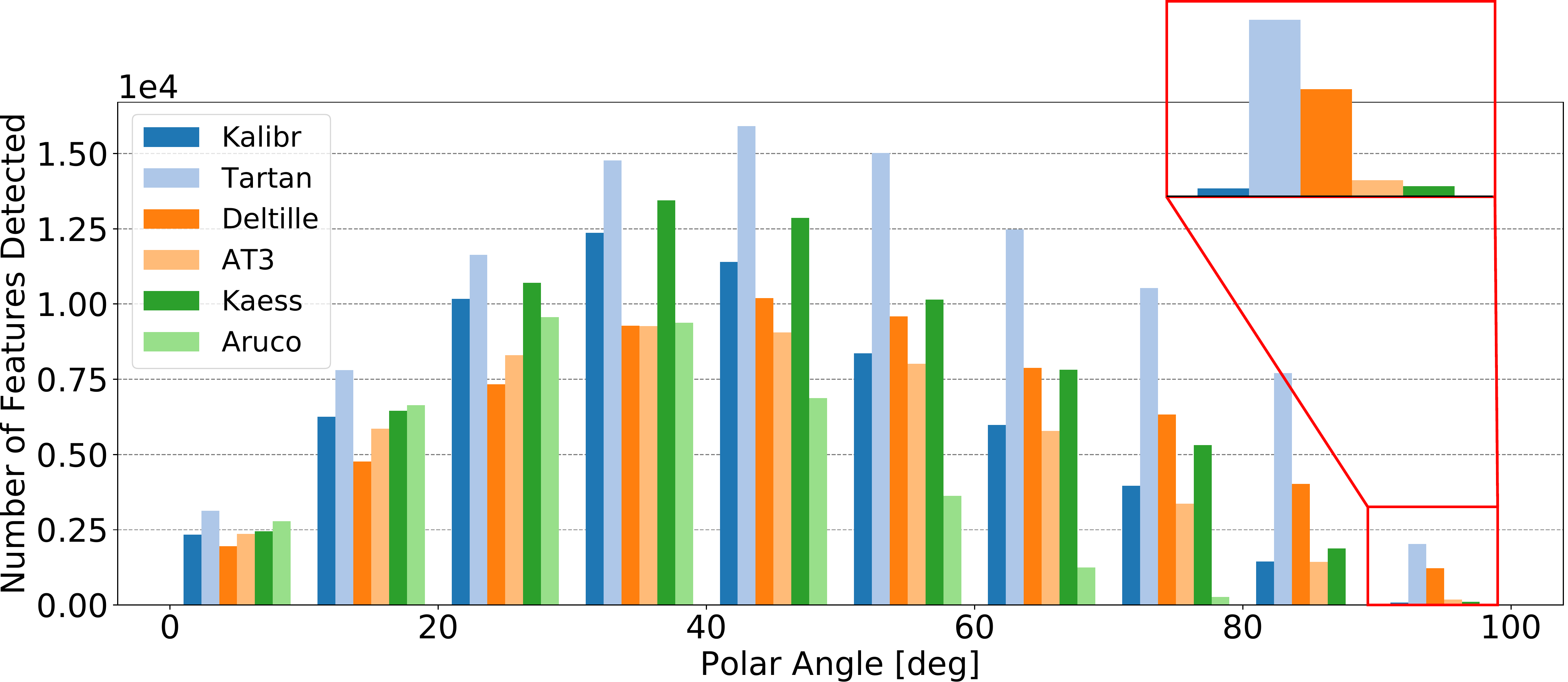}
  \caption{\apriltag{} features detected on the Lensagon BF5M. The results show that especially at the edge of the lens with polar angle approaching 100 degrees, \modelname{} detects significantly more features by reprojecting points from the target frame into the image frame. }
  \label{fig:features_polar_coords}
\end{figure}

While $\SI{5}{\percent}$-$\SI{10}{\percent}$ more features is a significant improvement, it does not resolve the issue of too little features at the edge of the lens. Reprojecting the features from the target frame into the image frame, as described in Section~\ref{sec:reprojection}, is therefore investigated as an alternative solution. For this experiment the dataset described in Section~\ref{sec:experiments_setup} is used, with 500 images for each evaluation. Figure~\ref{fig:features_polar_coords} shows the distribution of detected features by the various methods considered. \modelname{} is compared against other state-of-the-art \apriltag{} detectors, being: (1) Kalibr, (2) Deltille~\cite{8237833}, (3) \apriltag 3 (AT3)~\cite{5979561,wang2016apriltag}, a C++ implementation of AprilTags, originally written by Dr. Michael Kaess, and (4), the  ArUco tag detector~\cite{kallwies2020determining,romero2018speeded,garrido2016generation}. Table~\ref{tab:features_detected} shows the number of features detected using each method, whereas Table~\ref{tab:ord_6x6_distr} shows the number of features detected, sorted by polar angle.

The results show that \modelname{} is the most robust against lens distortion, recording far more features at the edge of the Lensagon BF5M fisheye lens. The results for GoPro (not shown in the figure) show that the numbers of detected features are roughly the same across all models. We attribute the reason to that it is a far less challenging lens to calibrate with its smaller FOV. The experiments that follow in later sections use reprojection and no undistortion.

\subsection{Subpixel Refinement}
In this section we compare two novel subpixel refinement strategies, and show that symmetry-based refinement is too unstable for grids of \apriltag{s}. The authors of~\cite{9156397} deployed low to medium distortion lenses and tested symmetry-based refinement on deltille grids and checkerboards.   Especially in areas of high distortion the refinements appear to be misaligned with the feature.
\begin{figure}[!htb]
    \centering
    \includegraphics[width=0.7\linewidth]{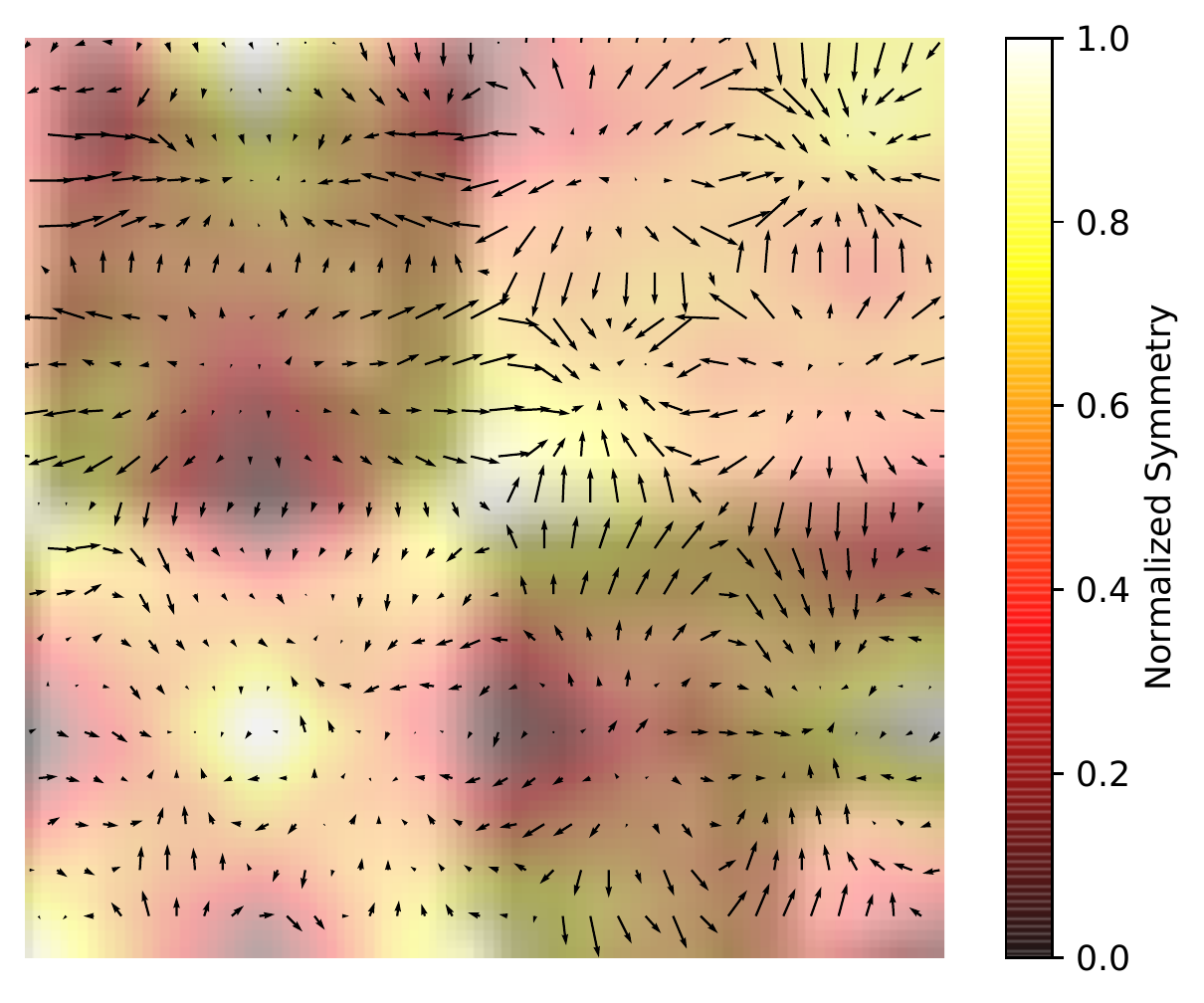}
    \caption{This figure shows the measure of symmetry as was proposed in Equation~\ref{eq:fe_symmetry}, on top of an \apriltag{} corner. The arrows are pointed in the direction of the symmetry gradient to show how an optimizer may behave locally. } 
    \label{fig:symmetry_refinement}
\end{figure}

Figure~\ref{fig:symmetry_refinement} shows why symmetry is an unsuitable metric for feature refinement with \apriltag{s}. The figure shows that numerous local optima exist, making a high-quality initial estimate a requirement. The initial estimate of the target position in the image frame is possibly several pixels off (as shown in Figure~\ref{fig:reprojection}), and may therefore converge to a local optimum. 
The resulting features, therefore, match the model better but do not appear to match the features in the real-world in high-distortion areas. We therefore hypothesize that adaptive \intextfunc{cornerSubPix()} yields more stable features. This hypothesis is studied in more detail in Section~\ref{subsec:human}.

\subsection{Comparison on Human-Annotated Features}
\label{subsec:human}
The closest thing to ground-truth data would be human-annotated feature locations. In this experiment, a subject with no access to the detections, annotated perceived feature locations through software. The observations were then compared against \modelname{} with adaptive \intextfunc{cornerSubPix()} and symmetry-based refinement. While a human is unable to detect features at a subpixel level, the experiment remains useful to quantify our observations. All annotated data are available through the project website. Table~\ref{tab:human} shows that \modelname{} outperforms Kalibr. Symmetry-based refinement outperforms adaptive \intextfunc{cornerSubPix()} in the GoPro dataset, but performs worse on the high-distortion Lensagon BF5M dataset. This shows that symmetry is an unstable metric for feature refinement, as was previously demonstrated in Figure~\ref{fig:symmetry_refinement}. For later experiments, we will exclusively use adaptive \intextfunc{cornerSubPix()} unless explicitly stated otherwise. 


\begin{table}[!htb]
\caption{\label{tab:human} The average detection error (px) in feature detection as compared to human-annotated data on 1,440 data points. The results show \modelname{} outperforms Kalibr significantly, and it shows symmetry-based refinement performs slightly better than \intextfunc{cornerSubPix()} on the GoPro dataset, but it performs worse on the high-distortion Lensagon BF5M fisheye lens dataset. This shows that symmetry is an unstable metric for feature refinement. }
\begin{tabular}{clll}
\hline
\multirow{2}{*}{Camera} & \multicolumn{1}{c}{\multirow{2}{*}{Kalibr}} & \multicolumn{2}{c}{TartanCalib}                   \\ \cline{3-4} 
                        & \multicolumn{1}{c}{}                        & Symmetry-Based & Adp. \intextfunc{cornerSubPix()} \\ \hline
Lensagon BF5M           & 1.038                                       & 0.861          & \textbf{0.815}                   \\
GoPro Hero 8            & 0.926                                       & \textbf{0.904} & 0.915                            \\ \hline
\end{tabular}
\end{table}

\begin{figure}[!htb]
    \centering
    \includegraphics[width=\linewidth]{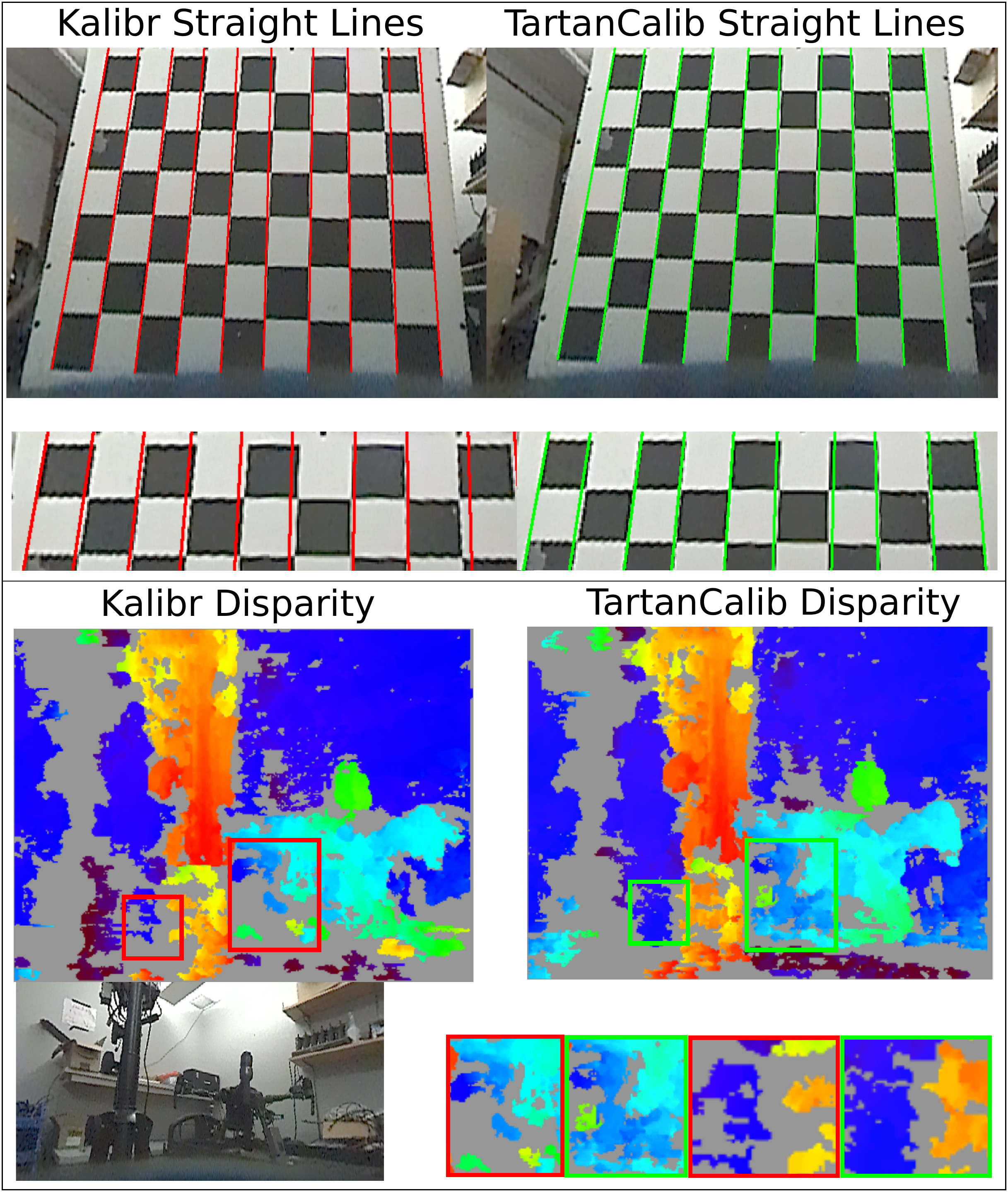}
    \caption{This Figure shows the results of an experiment for evaluating downstream performance when using pinhole reprojections at the edge of the Lensagon BF5M fisheye lens. We create a stereo pair of virtual pinhole images to show that (1) the images are less distorted using \modelname{}, and (2) the disparity maps generated using the stereo pair is less noisy. This result demonstrates better features detected by \modelname{} help improve downstream tasks.   }
    \label{fig:disparity}
\end{figure}

\subsection{Reprojection Error}
The most widely-used error metric for geometric camera calibration is reprojection error. Given an estimated target pose and camera model, it is possible to reproject the features back into the target frame and compare them against the detected corners (hence reprojection error). As aforementioned, this is not a particularly informative metric as the different methods output substantially different feature distributions (e.g., \modelname{} has far more features at the edge of the lens). Reprojection error is not only a function of the features, but also the ability of the camera model to fit the lens geometry. We perform a simple experiment to show that \modelname{} can improve reprojection error. Table~\ref{tab:reproj} shows a substantial improvement when using \modelname{}.

\begin{table}[]
\centering
\caption{\label{tab:reproj} Reprojection error evaluated on the Lensagon BF5M ultra-wide fisheye camera and GoPro Hero 8. DS-None: Double-Sphere model~\cite{usenko18double-sphere}, Omni-Radtan: Omnidirectional camera model with Radial-Tangential distortion~\cite{4409207}, and the Central Generic Model~\cite{9156397,8500466}} 
\begin{tabular}{llll}
\hline
Camera        & Model           & \begin{tabular}[c]{@{}l@{}}Kalibr \\ Reproj.\\ Err. {[}px{]}\end{tabular} & \begin{tabular}[c]{@{}l@{}}Tartan\\ Reproj.\\ Err. {[}px{]}\end{tabular} \\ \hline
Lensagon BF5M & Central Generic Model & 0.3297                                                                    & \textbf{0.2630}                                                          \\
GoPro Hero 8  & DS-None         & 1.0893                                                                    & \textbf{0.7981}                                                          \\
GoPro Hero 8  & Omni-Radtan     & 1.0869                                                                    & \textbf{0.7979}                                                          \\ \hline
\end{tabular}
\vspace{-5mm}
\end{table}

\subsection{Stereo Depth} \label{sec:stereo_depth}
Figure~\ref{fig:disparity} shows the results of an experiment that was performed to evaluate downstream performance when using pinhole reprojections at the edge of the Lensagon BF5M fisheye lens (see Figure~\ref{fig:drone}). We create a stereo pair of virtual pinhole images to (1) show that the images are less distorted using \modelname{}, and (2) the disparity maps generated using the stereo pair is less noisy. The results shows that the better features detected by \modelname{} help improve downstream tasks.

\section{Conclusion}
In this work, we have addressed the problem of geometric wide-angle camera calibration.  Previous methods lack features at the edge of the image region, where distortion is strongest. Our approach focuses on retrieving more features at the edge and improving their accuracy. We propose a novel calibration pipeline, \modelname{}, which iteratively improves camera models and leverages intermediate camera models to improve feature detections. Two novel subpixel refinement strategies are proposed, that leverage the intermediate model to achieve better accuracy. The results show that symmetry-based refinement is not a stable metric, and that a simple modification to \intextfunc{cornerSubPix()} yields the best results for high-distortion lenses.

Finally, the entire pipeline is implemented in an open-source easy-to-use toolbox. \modelname{} can be used as an augmentation to a state-of-the-art camera calibration toolbox, e.g. Kalibr, and improves the calibration effectiveness on wide-angle lenses.
With its iterative nature, \modelname{} does not require tedious hyperparameter tuning and typically only takes 2-3 times longer to run compared to the non-iterative baseline method. For wide-angle lenses \modelname{} delivers the highest feature coverage publicly available up to date.

\bibliographystyle{IEEEtran}
\bibliography{bibliography}  

\begin{thebibliography}{10}
\providecommand{\url}[1]{#1}
\csname url@samestyle\endcsname
\providecommand{\newblock}{\relax}
\providecommand{\bibinfo}[2]{#2}
\providecommand{\BIBentrySTDinterwordspacing}{\spaceskip=0pt\relax}
\providecommand{\BIBentryALTinterwordstretchfactor}{4}
\providecommand{\BIBentryALTinterwordspacing}{\spaceskip=\fontdimen2\font plus
\BIBentryALTinterwordstretchfactor\fontdimen3\font minus
  \fontdimen4\font\relax}
\providecommand{\BIBforeignlanguage}[2]{{%
\expandafter\ifx\csname l@#1\endcsname\relax
\typeout{** WARNING: IEEEtran.bst: No hyphenation pattern has been}%
\typeout{** loaded for the language `#1'. Using the pattern for}%
\typeout{** the default language instead.}%
\else
\language=\csname l@#1\endcsname
\fi
#2}}
\providecommand{\BIBdecl}{\relax}
\BIBdecl

\bibitem{9156397}
\BIBentryALTinterwordspacing
T.~Schops, V.~Larsson, M.~Pollefeys, and T.~Sattler, ``Why having 10,000
  parameters in your camera model is better than twelve,'' in \emph{2020
  IEEE/CVF Conference on Computer Vision and Pattern Recognition (CVPR)}.\hskip
  1em plus 0.5em minus 0.4em\relax Los Alamitos, CA, USA: IEEE Computer
  Society, jun 2020, pp. 2532--2541. [Online]. Available:
  \url{https://doi.ieeecomputersociety.org/10.1109/CVPR42600.2020.00261}
\BIBentrySTDinterwordspacing

\bibitem{usenko18double-sphere}
V.~Usenko, N.~Demmel, and D.~Cremers, ``The double sphere camera model,'' in
  \emph{Proc. of the Int. Conference on 3D Vision (3DV)}, September 2018.

\bibitem{Devernay2001StraightLH}
F.~Devernay and O.~D. Faugeras, ``Straight lines have to be straight,''
  \emph{Machine Vision and Applications}, vol.~13, pp. 14--24, 2001.

\bibitem{1642666}
J.~Kannala and S.~Brandt, ``A generic camera model and calibration method for
  conventional, wide-angle, and fish-eye lenses,'' \emph{IEEE Transactions on
  Pattern Analysis and Machine Intelligence}, vol.~28, no.~8, pp. 1335--1340,
  2006.

\bibitem{9711344}
\BIBentryALTinterwordspacing
Y.~Lochman, K.~Liepieshov, J.~Chen, M.~Perdoch, C.~Zach, and J.~Pritts,
  ``Babelcalib: A universal approach to calibrating central cameras,'' in
  \emph{2021 IEEE/CVF International Conference on Computer Vision
  (ICCV)}.\hskip 1em plus 0.5em minus 0.4em\relax Los Alamitos, CA, USA: IEEE
  Computer Society, oct 2021, pp. 15\,233--15\,242. [Online]. Available:
  \url{https://doi.ieeecomputersociety.org/10.1109/ICCV48922.2021.01497}
\BIBentrySTDinterwordspacing

\bibitem{Scaramuzza2006AFT}
D.~Scaramuzza, A.~Martinelli, and R.~Y. Siegwart, ``A flexible technique for
  accurate omnidirectional camera calibration and structure from motion,''
  \emph{Fourth IEEE International Conference on Computer Vision Systems
  (ICVS'06)}, pp. 45--45, 2006.

\bibitem{4059340}
D.~Scaramuzza, A.~Martinelli, and R.~Siegwart, ``A toolbox for easily
  calibrating omnidirectional cameras,'' in \emph{2006 IEEE/RSJ International
  Conference on Intelligent Robots and Systems}, 2006, pp. 5695--5701.

\bibitem{URBAN201572}
\BIBentryALTinterwordspacing
S.~Urban, J.~Leitloff, and S.~Hinz, ``Improved wide-angle, fisheye and
  omnidirectional camera calibration,'' \emph{ISPRS Journal of Photogrammetry
  and Remote Sensing}, vol. 108, pp. 72--79, 2015. [Online]. Available:
  \url{https://www.sciencedirect.com/science/article/pii/S0924271615001616}
\BIBentrySTDinterwordspacing

\bibitem{kalibr}
J.~Rehder, J.~Nikolic, T.~Schneider, T.~Hinzmann, and R.~Siegwart, ``Extending
  kalibr: Calibrating the extrinsics of multiple imus and of individual axes,''
  in \emph{2016 IEEE International Conference on Robotics and Automation
  (ICRA)}, 2016, pp. 4304--4311.

\bibitem{5979561}
E.~Olson, ``Apriltag: A robust and flexible visual fiducial system,'' in
  \emph{2011 IEEE International Conference on Robotics and Automation}, 2011,
  pp. 3400--3407.

\bibitem{8967787}
M.~Krogius, A.~Haggenmiller, and E.~Olson, ``Flexible layouts for fiducial
  tags,'' in \emph{2019 IEEE/RSJ International Conference on Intelligent Robots
  and Systems (IROS)}, 2019, pp. 1898--1903.

\bibitem{8500466}
J.~Beck and C.~Stiller, ``Generalized b-spline camera model,'' in \emph{2018
  IEEE Intelligent Vehicles Symposium (IV)}, 2018, pp. 2137--2142.

\bibitem{central_generic_2}
F.~Bergamasco, L.~Cosmo, A.~Gasparetto, A.~Albarelli, and A.~Torsello,
  ``Parameter-free lens distortion calibration of central cameras,'' in
  \emph{ICCV}, 10 2017, pp. 3867--3875.

\bibitem{937611}
M.~Grossberg and S.~Nayar, ``A general imaging model and a method for finding
  its parameters,'' in \emph{Proceedings Eighth IEEE International Conference
  on Computer Vision. ICCV 2001}, vol.~2, 2001, pp. 108--115 vol.2.

\bibitem{7516654}
S.~Ramalingam and P.~Sturm, ``A unifying model for camera calibration,''
  \emph{IEEE Transactions on Pattern Analysis and Machine Intelligence},
  vol.~39, no.~7, pp. 1309--1319, 2017.

\bibitem{10.1007/978-3-540-24671-8_1}
P.~Sturm and S.~Ramalingam, ``A generic concept for camera calibration,'' in
  \emph{Computer Vision - ECCV 2004}, T.~Pajdla and J.~Matas, Eds.\hskip 1em
  plus 0.5em minus 0.4em\relax Berlin, Heidelberg: Springer Berlin Heidelberg,
  2004, pp. 1--13.

\bibitem{888718}
Z.~Zhang, ``A flexible new technique for camera calibration,'' \emph{IEEE
  Transactions on Pattern Analysis and Machine Intelligence}, vol.~22, no.~11,
  pp. 1330--1334, 2000.

\bibitem{opencv}
\BIBentryALTinterwordspacing
G.~Bradski, ``\BIBforeignlanguage{English}{The opencv library},''
  \emph{\BIBforeignlanguage{English}{Dr.Dobb's Journal}}, vol.~25, no.~11, pp.
  120--125, 11 2000, copyright - Copyright Miller Freeman Inc. Nov 2000; Last
  updated - 2021-09-14; CODEN - DDJOEB. [Online]. Available:
  \url{https://www.proquest.com/trade-journals/opencv-library/docview/202684726/se-2}
\BIBentrySTDinterwordspacing

\bibitem{8237833}
H.~Ha, M.~Perdoch, H.~Alismail, I.~S. Kweon, and Y.~Sheikh, ``Deltille grids
  for geometric camera calibration,'' in \emph{2017 IEEE International
  Conference on Computer Vision (ICCV)}, 2017, pp. 5354--5362.

\bibitem{879788}
J.~Heikkila, ``Geometric camera calibration using circular control points,''
  \emph{IEEE Transactions on Pattern Analysis and Machine Intelligence},
  vol.~22, no.~10, pp. 1066--1077, 2000.

\bibitem{cornersubpix}
W.~Förstner and E.~Gülch, ``A fast operator for detection and precise
  location of distinct points, corners and centres of circular features.'' in
  \emph{ISPRS intercommission conference on fast processing of photogrammetric
  data}, 1987.

\bibitem{wang2016apriltag}
J.~Wang and E.~Olson, ``Apriltag 2: Efficient and robust fiducial detection,''
  in \emph{2016 IEEE/RSJ International Conference on Intelligent Robots and
  Systems (IROS)}.\hskip 1em plus 0.5em minus 0.4em\relax IEEE, 2016, pp.
  4193--4198.

\bibitem{kallwies2020determining}
J.~Kallwies, B.~Forkel, and H.-J. Wuensche, ``Determining and improving the
  localization accuracy of apriltag detection,'' in \emph{2020 IEEE
  International Conference on Robotics and Automation (ICRA)}.\hskip 1em plus
  0.5em minus 0.4em\relax IEEE, 2020, pp. 8288--8294.

\bibitem{romero2018speeded}
F.~J. Romero-Ramirez, R.~Mu{\~n}oz-Salinas, and R.~Medina-Carnicer, ``Speeded
  up detection of squared fiducial markers,'' \emph{Image and vision
  Computing}, vol.~76, pp. 38--47, 2018.

\bibitem{garrido2016generation}
S.~Garrido-Jurado, R.~Mu{\~n}oz-Salinas, F.~J. Madrid-Cuevas, and
  R.~Medina-Carnicer, ``Generation of fiducial marker dictionaries using mixed
  integer linear programming,'' \emph{Pattern recognition}, vol.~51, pp.
  481--491, 2016.

\bibitem{4409207}
C.~Toepfer and T.~Ehlgen, ``A unifying omnidirectional camera model and its
  applications,'' in \emph{2007 IEEE 11th International Conference on Computer
  Vision}, 2007, pp. 1--5.

\end{thebibliography}


\end{document}